\documentclass[11pt, a4paper, logo, copyright, nonumbering]{antgroup}
\usepackage[utf8]{inputenc}
\usepackage[square, numbers]{natbib}
\usepackage{dblfloatfix}
\usepackage{ulem}
\usepackage{caption}
\usepackage{dramatist}
\usepackage{xspace}
\usepackage{pifont}
\usepackage{multirow}

\usepackage{xltabular}
\usepackage{longtable}
\usepackage{hyperref}
\usepackage{amsfonts}
\usepackage{amsmath}
\usepackage{amssymb}
\usepackage{lineno}
\usepackage{adjustbox}

\usepackage[bottom]{footmisc}

\usepackage{CJKutf8}
\usepackage{subfigure}
\usepackage{setspace}

\usepackage{dsfont}
\usepackage{array}
\usepackage{tabularx}
\usepackage{subfigure}
\usepackage{xcolor}
\usepackage{tabularx}
\usepackage{booktabs}
\usepackage{subcaption}
\usepackage{graphicx}

\usepackage{lipsum}
\usepackage{multicol}
\usepackage{authblk}
\usepackage{wrapfig}
\usepackage[most,skins,theorems]{tcolorbox}
\usepackage{array}
\usepackage{pifont}
\definecolor{customgray}{RGB}{230,230,230}

\usepackage{graphicx}
\usepackage{multirow}
\usepackage{booktabs}
\usepackage[table]{xcolor}
\usepackage{hyperref}
\hypersetup{
	colorlinks=true,
	linkcolor=red,   
	urlcolor=blue,
}
\usepackage{array}
\usepackage{amssymb}
\usepackage{makecell}
\usepackage{graphicx}
\usepackage{adjustbox}
\usepackage{pifont}
\usepackage{subcaption}
\usepackage{caption}

\tcbset{
  aibox/.style={
    width=\linewidth,
    top=8pt,
    bottom=4pt,
    colback=gray!10!white,
    colframe=gray!50!black,
    colbacktitle=gray!70!black,
    enhanced,
    center,
    attach boxed title to top left={yshift=-0.1in,xshift=0.15in},
    boxed title style={boxrule=0pt,colframe=white,},
  }
}
\newtcolorbox{AIbox}[2][]{aibox,title=#2,#1}

\makeatletter
\def\@BTrule[#1]{%
  \ifx\longtable\undefined
    \let\@BTswitch\@BTnormal
  \else\ifx\hline\LT@hline
    \nobreak
    \let\@BTswitch\@BLTrule
  \else
     \let\@BTswitch\@BTnormal
  \fi\fi
  \global\@thisrulewidth=#1\relax
  \ifnum\@thisruleclass=\tw@\vskip\@aboverulesep\else
  \ifnum\@lastruleclass=\z@\vskip\@aboverulesep\else
  \ifnum\@lastruleclass=\@ne\vskip\doublerulesep\fi\fi\fi
  \@BTswitch}
\makeatother

\addto\extrasenglish{
}

 {\begin{list}{}%
         {\setlength{\leftmargin}{#1}}%
         \item[]%
 }
 {\end{list}}

\reportnumber{001} %

\renewcommand{\today}{}

\title{\centering Visual-Seeker: Towards Visual-Native Multimodal Agentic Search via Active Visual Reasoning}

\author{
Zhengbo Zhang$^{1,2,3, *}$,
Changtao Miao$^{3,*, \dag}$,
Jinbo Su$^{4,*}$,
Zhaowen Zhou$^{3,*, \dag}$,
Chunxia Zhang$^{5, \ddag}$,
Xukai Wang$^{2}$,
Ruiqi Liu$^{2}$,
Kaiyuan Zheng$^{3}$,
Jiansheng Cai$^{3}$,
Bo Zhang$^{3}$,
Zhe Li$^{3}$,

\vspace{-6pt}
Shiming Xiang$^{1,2}$,
Ying Yan$^{3, \ddag}$
\\
\vspace{-6pt}
$^{1}$ School of Artificial Intelligence, UCAS \hspace{2pt}
$^{2}$ Institute of Automation, CAS \\
$^{3}$ Ant Digital Technologies, Ant Group  \hspace{2pt}
$^{4}$ RUC \hspace{2pt}
$^{5}$ BIT
\vspace{-6pt}
}

\footnotetext[1]{Equal contribution.}
\footnotetext[2]{Project leader.}
\footnotetext[3]{Corresponding to cxzhang@bit.edu.cn, fuying.yy@antgroup.com}

\renewcommand{\phi}{\varphi}

\renewcommand{\epsilon}{\varepsilon}
\renewcommand{\imath}{\mathrm{i}}

\newlength{\restsubwidth}
\newlength{\restsubheight}
\newlength{\restsubmoreheight}
\newcommand{\rest}[2]{%
        \settowidth{\restsubwidth}{\ensuremath{#2}}
        \settoheight{\restsubheight}{\ensuremath{{}_{#2}}}
        \ensuremath{{#1\hskip 0.5pt}_{\vrule\kern2pt\parbox[b][%
        4pt][b]{\the\restsubwidth}{%
                        \ensuremath{{}_{#2}}}}}
        }

\begin{abstract}
Multimodal large language models (MLLMs) have demonstrated impressive capabilities in many visual tasks, but they often struggle with factual grounding when confronted with complex, open-world scenarios. 
While recent multimodal deep search agents attempt to address this issue by utilizing external tools, the visual-native search paradigm remains underexplored. Existing methods primarily rely on simple images with explicit semantics and text-only evidence trajectories, limiting the agent's ability to perform multi-hop, cross-modal reasoning and search.
To address these limitations, we propose \textbf{Visual-Seeker}, a visual-native multimodal deep search agent via active visual reasoning. Rather than treating vision as a static input, our agent actively attends to fine-grained visual details, dynamically harvests visual evidence throughout the search process. To unlock its visual-native potential, we design an active visual reasoning data pipeline and synthesize 5K high-quality multimodal trajectories for model training. Extensive experiments demonstrate the state-of-the-art performance across five challenging multimodal search benchmarks, even surpassing several proprietary models, validating robust visual-native reasoning and search in real-world web environments. The code and data can be accessed at: \href{https://github.com/ZhengboZhang/Visual-Seeker}{https://github.com/ZhengboZhang/Visual-Seeker}.
\end{abstract}

\begin{document}
\maketitle

\section{Introduction}

The rapid progress in large language models (LLMs) has spurred the development of autonomous search agents capable of multi-hop reasoning, tool use and web navigation \cite{search-r1, webdancer, browseragent}. However, these agent systems operate only on a text modality, making them difficult to handle visual queries and the visual information in the web environment. To bridge this gap, recent works have extended deep search agents to possess multimodal capabilities and equipped them with large multimodal language models (MLLMs) to handle image and text inputs \cite{mmsearch-r1, sensenova, skywork}. This improvement has driven the development of deep search agents for solving visual question answering (VQA) in the open world.

\newpage

\begin{wrapfigure}[27]{r}{0.58\textwidth}
  \centering
  \includegraphics[width=\linewidth]{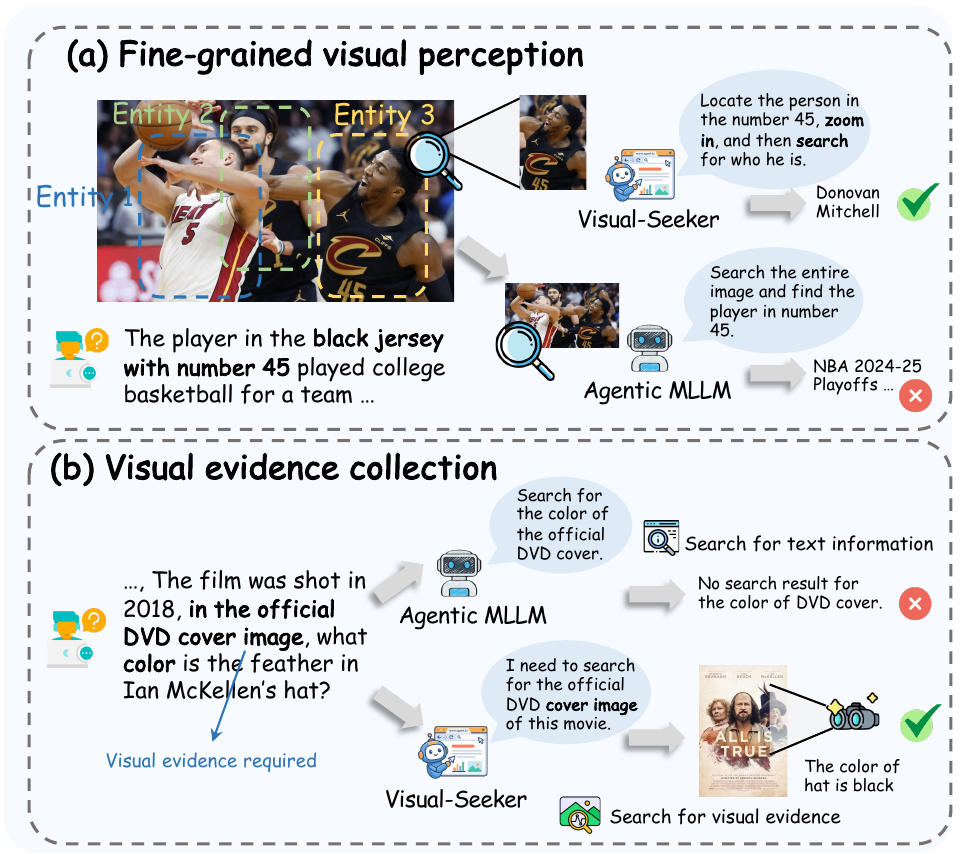}
  \caption{(a) The real-world image queries typically present complex, entity-rich visual content. The model must leverage its robust visual understanding to accurately identify target entities and iteratively refine the search process. (b) Multimodal Search Agent should actively aggregate multimodal evidence from web sources, synthesizing these diverse cues through cross-modal reasoning to generate comprehensive answers.}
  \label{fig1}
\end{wrapfigure}

But the visual-native capabilities of multimodal deep search agents remain largely unexplored. Visual information is only used as an adjunct modality to the query input, which limits the model's ability to handle scenarios with long-horizon visual reasoning.
As shown in Figure \ref{fig1}, first, the images that users query often come from the real world, and these images have complex backgrounds and multiple entities, which poses a challenge to the visual perception capabilities of multimodal search agents. However, the existing methods often construct their training data starting from images with entities containing explicit semantic information through web search, which hinders the training of visual perception capabilities \cite{webwatcher, mmds, mmsearch-r1}. Second, the real-world web environment contains rich visual information, therefore multimodal search agents should be capable of proactively collecting visual evidence. However, the most existing methods only insert visual queries into the prompt, without search trajectories that rely on visual evidence, which limits the coverage of multi-hop and multimodal scenarios \cite{sensenova, skywork, webwatcher}.

To address these limitations, we propose a visual-native multimodal deep search agent, \textbf{Visual-Seeker} that bridges the gap between passive visual perception and active cross-modal search in open-world environments. 
As illustrated in Figure \ref{fig1}, in contrast to existing methods, our agent is designed to enhance the visual-native capability in search trajectory through active visual reasoning. Specifically, our model can perceive fine-grained properties in complex real-world images containing multiple interconnected entities, dynamically perform multi-hop reasoning, and proactively collect and integrate visual evidence during the search process. 
To cultivate these visual-native capabilities, we design an \textbf{Active Visual Reasoning} data synthesis pipeline with three stages. First, we extract the names and visual descriptions of seed entities from real-world images with visual reasoning processes,thereby obtaining target entities in complex images. This drives the model to perform fine-grained visual perception and search of specific regions to obtain semantic information about the target entities. We then use a dual-strategy random walk to expand the depth and breadth of the search trajectory. Finally, we use search engines to obtain information-rich images of the entities and merge queries containing visual evidence. Based on this data pipeline, we synthesize 5K high-quality multimodal search trajectories and use them for training Visual-Seeker.
Extensive experiments across five challenging benchmarks demonstrate that our agent achieves the state-of-the-art performance, significantly outperforming existing open-source and proprietary models.

The key contributions of this work can be summarized as follows:

\begin{itemize}
    \item We propose a multimodal deep search agent, Visual-Seeker , which combines visual-native capabilities with search. It can perform visual understanding of images in complex multi-entity scenes and proactively collect visual evidence for cross-modal search.
    \item We design a Active Visual Reasoning data synthesis pipeline that enables models to develop fine-grained visual perception and the ability to actively collect visual evidence during deep search. 5K high-quality trajectories generated from the data pipeline are used for model training.
    \item Our agent achieves the state-of-the-art performance on five multimodal search benchmarks and outperforms several proprietary models.
\end{itemize}

\section{Related Works}

\subsection{Text-only Deep Search Agent}

The pre-trained knowledge of large language models is time-truncated, and the retrieval-augmented generation (RAG) \cite{rag1, rag2, xue2022multi} method has the limitation of pre-building a knowledge base. Text-only deep search agents aim to overcome this restriction by leveraging external tools to search in real-world environments \cite{search-r1, webdancer}. Text-only deep search agents transform complex information retrieval processes into iterative loops of reasoning and tool calls \cite{browseragent, tongyidr, websailor}. This paradigm empowers large language models to autonomously generate search queries and browse web pages. However, such text-based search agents are restricted by textual queries and document retrieval, thereby lacking the capacity to interpret or leverage multimodal sensory data from real-world scenarios.

\subsection{Multimodal Deep Search Agent}

The development of multimodal models has driven researchers to explore multimodal deep search agent. Early studies \cite{mmsearch-r1, webwatcher} equip the agent with reverse image search tools to obtain the semantic information of the entire image from the visual input, and fuse textual QA with entity-visual queries to generate fine-tuning data, enabling the agent to retrieve images and perform multi-turn reasoning. Recent works \cite{deepmmsearch, skywork, sensenova, deepeyes} have introduced the image cropping tool that leverages the model's visual grounding capabilities to retrieve target entities within images, mitigating the interference of background noise.

While the ability of agents to interact with tools and perform multi-turn reasoning is crucial, visual reasoning capabilities and the ability to proactively gather visual evidence are also indispensable in multi-turn search for solving complex problems \cite{mmbc, visbrowse}. However, the existing methods still have limitations in constructing fine-tuning data for multimodal search agent: lack of visual queries for complex, multi-entity images that closely resemble the real world; lack of incorporation of visual evidence into the necessary path of search.

\section{Method}

\subsection{Active Visual Reasoning Data Pipeline}
\label{data}

\begin{figure}[t]
    \centering
    \includegraphics[width=1\linewidth]{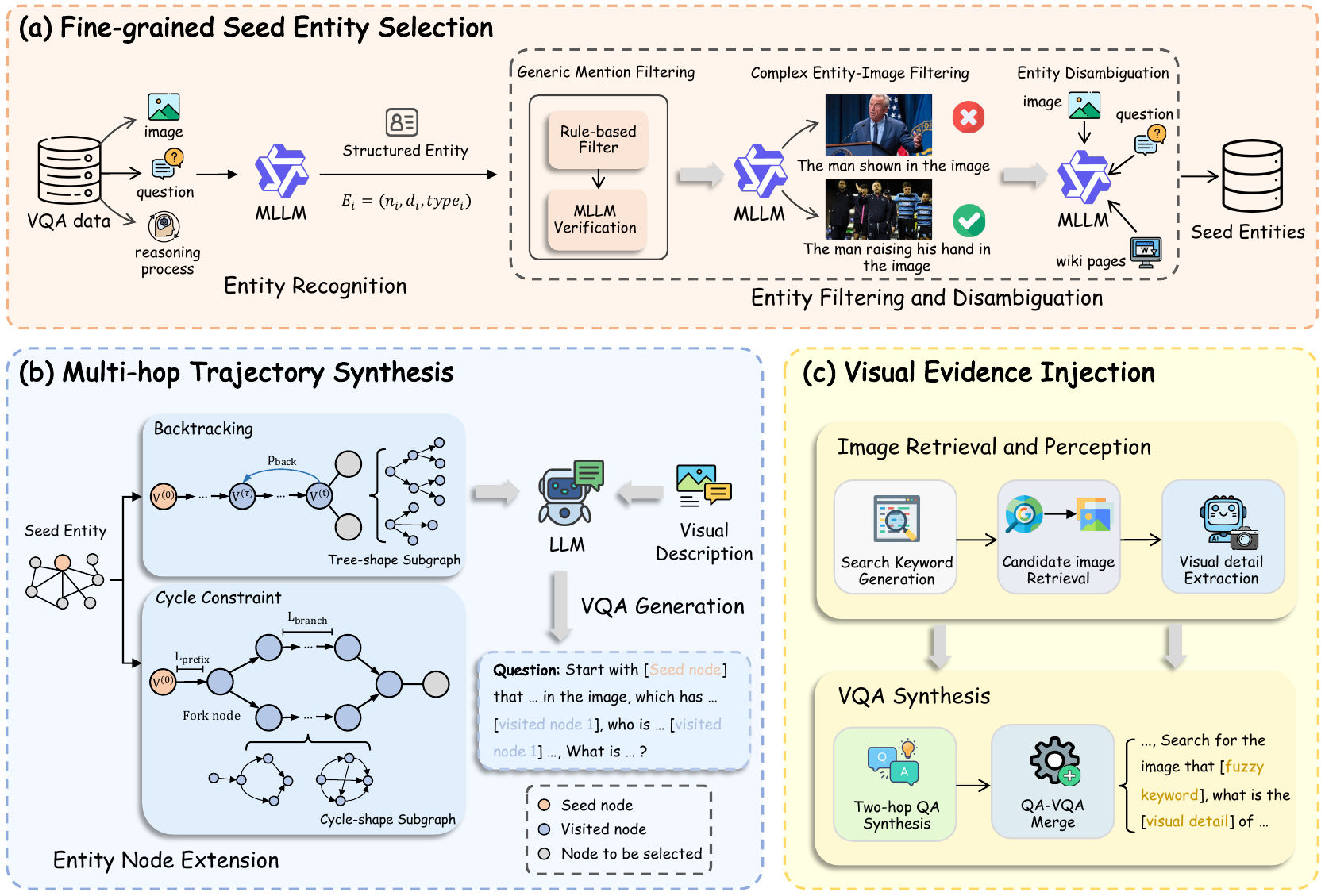}
    \caption{\textbf{Active Visual Reasoning Data Pipeline.} This pipeline synthesizes complex visual queries by extracting entity information from multi-entity images, expands the search depth through random walks on a knowledge graph, and ultimately inserts visual evidence into the search trajectory. (a) Multi-entity visual description extraction and entity name filtering and disambiguation. (b) Dual-strategy random walk expands search depth. (c) Injection of visual evidence to force visual search and reasoning.}
    \label{fig2}
\end{figure}

To construct high-quality training data for multimodal deep search agent, we propose an active visual reasoning data synthesis pipeline for visual-native search task. As illustrated in Figure \ref{fig2}, our pipeline generates multi-hop multimodal deep search trajectories, starting from complex entity-centric queries and strategically injecting visual evidence to activate the visual-native reasoning capabilities of MLLMs.

\subsubsection{Fine-grained Seed Entity Selection}
\label{ses}

Accurately obtaining the target seed entity to be searched in multi-entity images is crucial for activating the fine-grained visual perception ability of the model. LiveVQA \cite{livevqa} is a large-scale dataset featuring time-sensitive visual questions and multi-entity images. It also provides reasoning process for each VQA instance, enabling the extraction of multiple seed entities and their corresponding captions from complex real-world images. 

\paragraph{Entity Recognition.} Unlike conventional named entity recognition (NER) that operates solely on text, our approach leverages both visual and textual cues to identify entities that are visually grounded and relevant to the reasoning chain. Given a query image $I$, question $Q$ and reasoning process $R$ sampled from LiveVQA, we employ a MLLM with in-context learning to perform joint entity extraction. The MLLM is instructed to output a structured list of entities, where each entity $E_i$ is represented as:

\begin{equation}
    E_i = (n_i, d_i, type_i)
\end{equation}

\noindent where $n_i$ is the exact name of the entity and $d_i$ is a concise textual description of the entity's state in the image, such as ``The man wearing a pink shirt in the picture''. And $type_i$ is the category of the entity, such as person, location, organization, etc. 

\paragraph{Entity Filtering and Disambiguation.} Not all extracted entities are suitable for grounding search queries. We apply a three-step filtering strategy to ensure entity quality.

\textit{Step 1: Generic Mention Filtering.} We filter out entities that lack specific identifiers, such as generic noun phrases or pronouns ``the man'', ``a building''. Formally, an entity $E_i$ is retained only if its name $n_i$ is a proper noun or a uniquely identifiable common noun within the given context. This is implemented through a rule-based filter combined with MLLM verification.

\textit{Step 2: Complex Entity-Image Filtering.} Complex multi-entity images are key to increasing the difficulty of the reasoning questions, so we need to filter out images with obvious entity semantics. Specifically, we construct the context from the query image $I$ and entity description $d_i$, asking MLLM to determine whether the semantic information of entities in an image can be easily obtained, such as a photograph of single person with the entity description stating ``The main character in the picture''.

\textit{Step 3: Entity Disambiguation.} For polysemous entities, such as ``Apple'' which refers to both a company and a fruit, we perform context-aware disambiguation. We construct a disambiguation prompt that provides the query image $I$, question $Q$, and surrounding entities as context, asking the MLLM to select the correct sense from candidate Wikipedia disambiguation pages.

After entity recognition, filtering, and disambiguation, we obtained query images with complex semantic entities, and 2K entities from it as seed entities for multi-hop VQA synthesis.

\subsubsection{Multi-hop Trajectory Synthesis}

Given the seed entities, we synthesize multi-hop reasoning questions by randomly walking on an offline Wikipedia knowledge graph. The core objective is to generate diverse and non-linear search trajectories that drive the search agent in calling tools and collecting evidence from multiple sources. 

\paragraph{Entity Node Extension}

We construct an offline knowledge graph $\mathcal{G} = (\mathcal{V}, \mathcal{E})$  from the Wikipedia and perform random walks starting from the node of seed entity $V^{(0)}$, recursively following links from each entity node to simulate human browsing behavior. A naive depth-first random walk on $\mathcal{G}$ produces linear reasoning chains, which are insufficient for training robust search agents. So, we introduce two strategies to extend the topology of the subgraph.

\textit{Strategy 1: Backtracking.} To simulate the realistic behavior of search agents revisiting previous assumptions, we use backtracking strategy. At step $t$, with probability $p_{back} \in (0,1)$, the walker jumps back to a previously visited entity $V^{(\tau)}$, where $\tau < t$, chosen from the walk history $\mathcal{H}_t = \{V^{(0)}, ..., V^{(t)}\}$:

\begin{equation}
    V^{(t+1)} = 
    \begin{cases}
    V^{(\tau)} \sim \mathcal{H}_t & \text{with } p_{\text{back}}, \\
    V' \sim \mathcal{N}(V^{(t)}) & \text{otherwise}.
    \end{cases}
\end{equation}

\noindent where $\mathcal{N}(V^{(t)})$ is the set $V^{(t)}$'s neighbor nodes. This strategy creates tree-shape structures within the trajectory, forcing the agent to manage and compare multiple reasoning branches.

\textit{Strategy 2: Cycle Constraint}
Starting from the seed entity, the walker first traverses a shared prefix of length $L_{prefix}$ to reach a fork node. At the fork, two disjoint branches are spawned and each extending independently for $L_{branch}$ steps with mutual node exclusion to ensure path diversity. The branches then converge toward a common node through iterative expansion of their frontier sets, with a maximum of $T$ attempts to discover a valid convergence point. A BFS verification confirms the existence of two node-disjoint paths between fork and convergence, ensuring the strict cycle topology.

\paragraph{QA Generation}

We sample connected subgraphs $\mathcal{G}_s \in \mathcal{G}$ from the topology obtained by entity node extension. With the relationships between entities in the subgraph as the context for reasoning, we prompt a large language model to generate text questions that conform to the subgraph constraints.

\subsubsection{Visual Evidence Injection}

While the knowledge-graph-based QA synthesis equips the agent with structured factual reasoning, it neglects a critical capability in multimodal deep search: actively acquiring and interpreting visual evidence from the web pages. To address this gap, we introduce visual evidence into the reasoning process of synthetic VQA, forcing the agent to call tools to search for images and perform pixel-level reasoning.

For each VQA instance, we locate the name of its answer entity and its attribute description from the wiki page. Then we prompt a LLM to generate a search keyword, and use a search engine to retrieve a set of candidate images. This allows for the retrieval of images containing richer visual information, activating the model's visual understanding capabilities.

For each candidate image, we use an MLLM to extract visual details that are not derivable only from textual descriptions. The MLLM is prompted to identify fine-grained attributes in the image, such as color patterns and spatial layouts. Then, we use fuzzy search keywords as question and visual details as answer to synthesize a two-hop QA pair, driving the model to invoke tools for text-to-image retrieval and visual perception. Finally, we merge the extended two-hop QA into the original VQA. The examples of our synthesized data are shown in the Appendix \ref{ap:data}.

\subsection{Visual-Seeker}

Based on the VQA dataset synthesized in our data pipeline, we train the Visual-Seeker using a carefully crafted supervised fine-tuning method without relying on costly reinforcement learning. We employ the teacher model to generate multimodal search trajectories based on our agentic workflow and perform SFT as cold start.

\begin{figure}[t]
    \centering
    \includegraphics[width=1\linewidth]{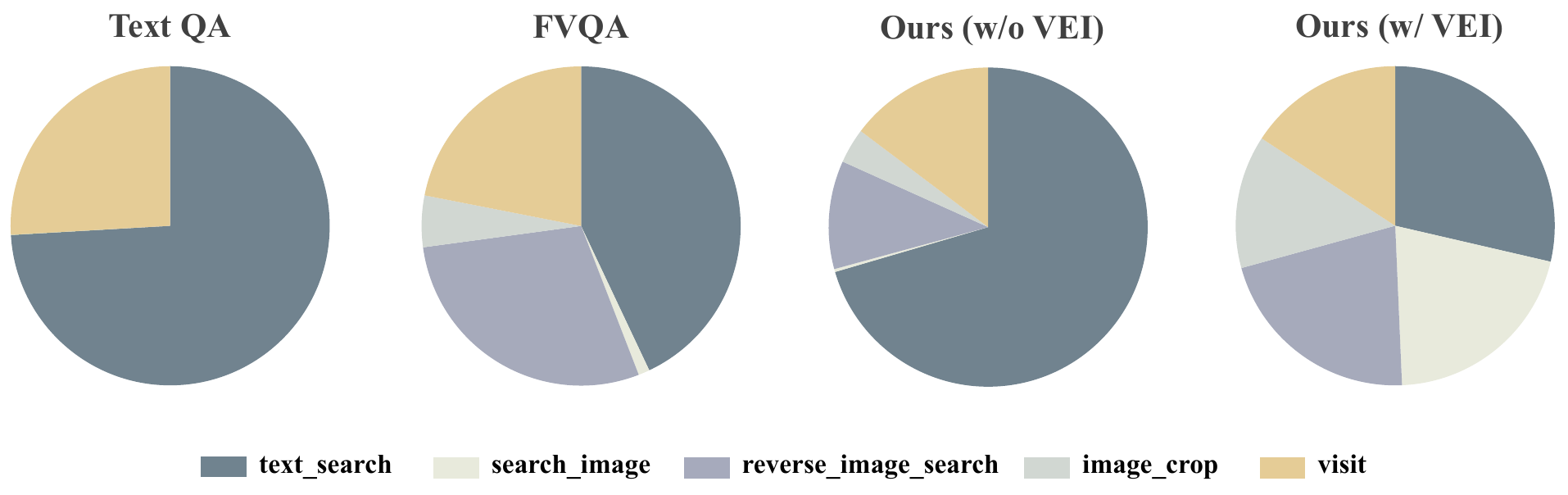}
    \caption{\textbf{Tool call distribution of training trajectories.} Our synthesized queries require more \texttt{search\_image} tool calls and more balanced tool call pattern. The VEI shorts for Visual Evidence Injection.}
    \label{fig4}
\end{figure}

\subsubsection{Workflow and Tools}

We provide the agent with five external tools that it can choose to use, including: (1) \texttt{text\_search}, powered by SerperAPI \footnote{https://serper.dev}, it can retrieve relevant webpage titles and URLs based on natural language queries; (2) \texttt{reverse\_image\_search} retrieves images related to the input image and return the webpage titles and URLs; (3) \texttt{search\_image} retrieves images related to the textual input for active visual evidence collection; (4) \texttt{visit} uses JinaAPI \footnote{https://jina.ai} to capture webpage content and submit it to an summary model for summarization, and we use Qwen3-30B-A3B-Instruct as summary model; (5) \texttt{image\_crop} crops the image based on the input coordinates for fine-grained visual reasoning.

The multimodal search agent operates in a ReAct-style loop. Given a text query $Q$ and optional images $I$, the context $C_i$ in the $i$-th interaction round is denoted as:

\begin{equation}
    C_i = (Q, I, R_1, A_1, \mathcal{O}_1, ..., R_i, A_i, \mathcal{O}_i)
\end{equation}

\noindent Where $R$ is the reasoning content of model, $A$ represents the tool call parameters within <tool\_call> blocks, and $\mathcal{O}$ represents the results returned by the tools. The loop terminates when the model outputs an answer or reaches the maximum number of tool interaction rounds, which we set to 15. 

\subsubsection{Supervised Fine-Tuning}

We first perform SFT for a cold start, teaching the base model the essential capabilities required for a multimodal search agent. Following the data pipeline described in Section \ref{data}, we ultimately collected 3K VQA data without visual evidence injection and 800 VQA data with visual evidence injection. Based on our agentic workflow, we use Claude-4.6-Opus \cite{claude} as the teacher model and generate multimodal deep search trajectories. In addition, we synthesize text-only QA instances and generated 500 trajectories. And we sample VQA instances from the FVQA \cite{mmsearch-r1} training set and generate 700 multimodal trajectories. Finally, we mix all the above 5K data and train the model using cross-entropy loss. The goal of SFT is to guide the model to learn a pattern of multi-turn interaction with the tools, and to actively collect and organize visual and textual evidence during the search process.

The distribution of tool call counts in the trajectories of various data sources is shown in Figure \ref{fig4}. The high percentage of \texttt{text\_search} tool calls based on the data constructed using our data pipeline indicates that our queries require multi-hop search. After introducing visual evidence injection, the proportion of triggers for \texttt{search\_image} tools in the agent trajectory increases significantly. This indicates that our synthesized queries encourage the model to collect multimodal evidence from web pages for deeper search process.

\section{Experiments}

\subsection{Experimental Setup}

\paragraph{Implementation details.}
During the SFT stage, we use Ms-Swift \cite{ms-swift} as the training framework to full fine-tuning of the model. The model is trained over 3 epochs with a batch size of 8, and a learning rate of 2e-6. 
We use Qwen3-VL-8B-Instruct \cite{qwen3} as the base model and the training stage is conducted on 8 NVIDIA A100 GPUs.

\paragraph{Evaluation benchmarks.}
We evaluate our model on five challenging multimodal agentic search benchmarks: MMSearch \cite{,,mmsearch}, MMSearch-Plus \cite{mmsearchp}, BrowseComp-VL \cite{webwatcher}, MM-BrowseComp \cite{mmbc} and VisBrowse-Bench \cite{visbrowse}. For MMsearch-Plus, based on previous work \cite{vdr, points}, we only use single-image samples.

\paragraph{Baselines.}
We use three types of methods for the baseline models: direct answer, agentic workflow and multimodal deep search agent. The evaluated models include proprietary and open-source multimodal models: GPT-5 \cite{gpt5}, Gemini-2.5 series \cite{gemini}, Claude-4-Sonnet \cite{claude}, Qwen3-VL-8B-Instruct \cite{qwen3} and various multimodal deep search agents.

\begin{itemize}
    \item \textbf{Direct Answer:} Models answer the question relying on internal parametric knowledge, without external tool access.
    \item \textbf{Agentic Workflow:} Models can use all the tools in our agent framework to collect visual and textual evidence.
    \item \textbf{Multimodal Deep Search Agent:} We compare the performance with existing multimodal search agents.
\end{itemize}

\paragraph{Evaluation metrics.} We use accuracy (\%) as the metric to evaluate the model's performance on five benchmarks. Qwen3-235B-A22B-Instruct is employed as a judge model using the LLM-as-Judge method to evaluate answer correctness against ground truth.. Details of the prompts are shown in the Appendix \ref{ap:prompt}.

\begin{table}[!ht]
\centering
\caption{The performance comparison between our model and other methods across five challenging benchmarks. MMSearch+ shorts for MMSearch-Plus, BC-VL shorts for BrowseComp-VL and MM-BC shorts for MM-BrowseComp. The \textbf{bold numbers} represent the best accuracy in each benchmark. The \textcolor{green!70!black}{$\Delta$} represents an improvement in our model compared to the base model in agentic workflow.}
\label{tab:main}
\resizebox{\linewidth}{!}{ 
\begin{tabular}{lcccccc}
\toprule
\textbf{Model} & \textbf{MMSearch} & \textbf{MMSearch+} & \textbf{BC-VL} & \textbf{MM-BC} & \textbf{VisBrowse} & \textbf{Avg.} \\
\midrule
\rowcolor{gray!20} \multicolumn{7}{c}{\textbf{Direct Answer}} \\
\midrule
\multicolumn{1}{l|}{GPT-5} & 33.3 & \textbf{19.1} & \textbf{47.2} & \textbf{10.3} & \textbf{26.0} & \textbf{27.2} \\
\multicolumn{1}{l|}{Gemini-2.5-Flash} & 30.4 & 8.1 & 37.1 & 5.4 & 16.0 & 19.4 \\
\multicolumn{1}{l|}{Gemini-2.5-Pro} & \textbf{39.8} & 14.5 & 43.1 & \textbf{10.3} & 19.5 & 25.4 \\
\multicolumn{1}{l|}{Claude-4-Sonnet} & 18.7 & 4.0 & 29.3 & 5.3 & 8.3 & 13.1 \\
\multicolumn{1}{l|}{Qwen3-VL-8B-Instruct} & 15.2 & 3.2 & 25.1 & 4.9 & 10.7 & 11.8 \\

\midrule
\rowcolor{gray!20} \multicolumn{7}{c}{\textbf{Agentic Workflow}} \\
\midrule
\multicolumn{1}{l|}{GPT-5} & 65.7 & \textbf{34.5} & \textbf{49.1} & 9.4 & \textbf{29.0} & \textbf{37.5} \\
\multicolumn{1}{l|}{Gemini-2.5-Flash} & 67.0 & 23.3 & 43.1 & 10.4 & 20.7 & 32.9 \\
\multicolumn{1}{l|}{Gemini-2.5-Pro} & 67.8 & 30.9 & 45.2 & \textbf{14.8} & 26.6 & 37.1 \\
\multicolumn{1}{l|}{Claude-4-Sonnet} & \textbf{70.3} & 20.9 & 44.1 & 6.3 & 19.5 & 32.2 \\
\multicolumn{1}{l|}{Qwen3-VL-8B-Instruct} & 53.8 & 10.9 & 28.4 & 6.7 & 15.4 & 23.0 \\

\midrule
\rowcolor{gray!20} \multicolumn{7}{c}{\textbf{Multimodal Deep Search Agent}} \\
\midrule
\multicolumn{1}{l|}{MMSearch-R1-7B \cite{mmsearch-r1}} & 53.8 & - & - & - & - & - \\
\multicolumn{1}{l|}{WebWatcher-7B \cite{webwatcher}} & 49.1 & - & 21.2 & - & - & - \\
\multicolumn{1}{l|}{WebWatcher-32B \cite{webwatcher}} & 55.3 & - & 27.0 & - & - & - \\
\multicolumn{1}{l|}{DeepEyesV2-7B \cite{deepeyes}} & 63.7 & - & - & - & - & - \\
\multicolumn{1}{l|}{Skywork-R1V4-30B-A3B \cite{skywork}} & 66.1 & - & 38.4 & - & - & - \\
\multicolumn{1}{l|}{SenseNova-MARS-8B \cite{sensenova}} & 67.8 & - & - & - & - & - \\
\multicolumn{1}{l|}{Vision-DeepResearch-8B \cite{vdr}} & 69.6 & 20.4 & 42.6 & - & - &  \\
\multicolumn{1}{l|}{VSearcher-8B \cite{vsearcher}} & 47.2 & - & 30.8 & 6.2 & - & - \\
\multicolumn{1}{l|}{MM-DeepResearch-8B \cite{mmds}} & 67.8 & - & 37.9 & - & - & - \\
\multicolumn{1}{l|}{MM-DeepResearch-32B \cite{mmds}} & 69.0 & - & 43.0 & - & - & - \\
\multicolumn{1}{l|}{Points-Seeker-8B \cite{points}} & 70.8 & 25.2 & 44.4 & - & - & - \\
\multicolumn{1}{l|}{OpenSearch-VL-8B \cite{opensearch}} & 64.5 & - & 37.6 & - & - & - \\
\midrule
\rowcolor{green!10} \multicolumn{1}{l|}{\textbf{Visual-Seeker (Ours)}} & \textbf{72.2} & \textbf{27.3} & \textbf{47.6} & \textbf{16.1} & \textbf{34.7} & \textbf{39.6} \\
\multicolumn{1}{l|}{\textcolor{green!70!black}{$\Delta$ Qwen3-VL-8B-Instruct (Agentic)}} & \textcolor{green!70!black}{\textbf{+18.4}} & \textcolor{green!70!black}{\textbf{+16.4}} & \textcolor{green!70!black}{\textbf{+19.2}} & \textcolor{green!70!black}{\textbf{+9.4}} & \textcolor{green!70!black}{\textbf{+19.3}} & \textcolor{green!70!black}{\textbf{+16.6}} \\

\bottomrule
\end{tabular}
}
\end{table}

\subsection{Main Results}

As shown in Table \ref{tab:main}, we evaluate proprietary MLLMs and multimodal search agents under three methods. Most models performe poorly in the direct answer approach, which is related to the limited pre-trained knowledge of the models; for example, Claude-4-Sonnet only achieves an average score of 13.1 on five Benchmarks. After integrating with our agent workflow, all models show significant performance improvements, with Claude-4-Sonnet achieving a 145.8\% increase. This demonstrates the robustness of our workflow and its applicability to various models. 

Our agent achieves an average accuracy of 39.6\% across five benchmarks, outperforming all current multimodal deep search agents and even competing with some proprietary models. Compared to Qwen3-VL-8B-Instruct (Agentic) baseline, our model achieves nearly double the performance on every benchmark. 

In MMSearch-Plus, which features complex multi-entity image queries, our model demonstrates strong competitiveness, proving a significant improvement in our visual understanding capabilities. In MM-BrowseComp and VisBrowse-Bench, where visual evidence is required during the search process, our model outperforms even the proprietary models GPT-5 and Gemini-2.5-Pro.

\subsection{Ablation Study and Analysis}

\paragraph{Data Ablation.} To validate the effectiveness of our data synthesis pipeline, we perform ablation analysis on data from different sources and modalities. Specifically, we incrementally add four types of data to the training set to train the base model using SFT. As shown in Table \ref{tab:data}, training the model using open-source multimodal query and text query trajectories allows it to learn tool calls and reasoning patterns, resulting in a slight performance improvement. The multimodal trajectories synthesized based on our data pipeline brings significant benefits to the performance of multi-hop search. For benchmark MMSearch-Plus requiring fine-grained visual perception, our data obtains a 17.2\% performance improvement. After training with data infused with visual evidence, the model's ability to search for visual information improved, resulting in significant improvements on MM-BrowseComp and VisBrowse.

\begin{table}[!ht]
\centering
\caption{Ablation results of training data. Four types of data, including FVQA, QA generated by our data pipeline, VQA without visual evidence injection and VQA with visual evidence injection.}
\label{tab:data}

\begin{tabular}{lcccc}
\toprule
\textbf{Data} & \textbf{MMS+} & \textbf{MM-BC} & \textbf{Vis} & \textbf{Avg.} \\
\midrule
\multicolumn{1}{l|}{Qwen3-VL-8B-Instruct} & 10.9 & 6.7 & 15.4 & 11.0 \\
\midrule
\multicolumn{1}{l|}{+ FVQA + QA traj.} & 20.9 & 6.3 & 10.7 & 12.6 \\
\multicolumn{1}{l|}{+ wo/ VEI VQA traj.} & 24.5 & 11.1 & 20.1 & 18.6 \\
\multicolumn{1}{l|}{+ w/ VEI VQA traj.} & \textbf{27.3} & \textbf{16.1} & \textbf{34.7} & \textbf{26.0} \\

\bottomrule
\end{tabular}

\end{table}

\paragraph{Tool Ablation.}

Our approach relies on visual reasoning and visual evidence search to achieve visual-native search, with two core tools \texttt{image\_crop} and \texttt{search\_image}. To verify the effectiveness of the two tools, we remove both tools respectively, and the ablation results are shown in Table \ref{tab:tool}, removing either tool will significantly reduce the model's performance across all benchmarks. The largest decline after removing the \texttt{image\_crop} tool occurs on the VisBrowse benchmark, indicating that image queries on this benchmark have multiple complex entities and it is difficult to obtain the semantics of the target entity through entire image search. The largest decline after removing the \texttt{search\_image} tool also occurs on the VisBrowse benchmark, indicating that this benchmark needs to integrate visual evidence to arrive at the correct answer in the search trajectory. After removing both two tools, the model's ability of active visual grounding and visual evidence collection degrades.

\begin{table}[!ht]
\centering
\caption{Ablation results of tools. `w/o IC' represents removing the \texttt{image\_crop} tool and `w/o SI' represents removing the \texttt{search\_image} tool. The \textcolor{red!80!black}{$\Delta$} represents the performance reduction compared to the model equipped with full tools.}
\label{tab:tool}
\begin{tabular}{lcccc}
\toprule
\textbf{Tools} & \textbf{MMS+} & \textbf{MM-BC} & \textbf{Vis} & \textbf{Avg.} \\
\midrule
\multicolumn{1}{l|}{Visual-Seeker} & 27.3 & 16.1 & 34.7 & 26.0 \\
\midrule
\multicolumn{1}{l|}{w/o IC} & 23.7 & 12.5 & 25.1 & 20.4 \\
\multicolumn{1}{l|}{\textcolor{red!80!black}{$\Delta$}} & \textcolor{red!80!black}{-3.6} & \textcolor{red!80!black}{-3.6} & \textcolor{red!80!black}{-9.6} & \textcolor{red!80!black}{-5.6} \\
\multicolumn{1}{l|}{w/o SI} & 22.7 & 11.7 & 20.1 & 18.2 \\
\multicolumn{1}{l|}{\textcolor{red!80!black}{$\Delta$}} & \textcolor{red!80!black}{-4.6} & \textcolor{red!80!black}{-4.4} & \textcolor{red!80!black}{-14.6} & \textcolor{red!80!black}{-7.8} \\
\multicolumn{1}{l|}{w/o IC \& SI} & 21.5 & 9.9 & 19.9 & 17.1 \\
\multicolumn{1}{l|}{\textcolor{red!80!black}{$\Delta$}} & \textcolor{red!80!black}{-5.8} & \textcolor{red!80!black}{-6.2} & \textcolor{red!80!black}{-14.8} & \textcolor{red!80!black}{-8.9} \\

\bottomrule
\end{tabular}
\end{table}

\paragraph{Analysis of Tool Usage.}

\begin{figure}[htbp]
    \centering
    \begin{minipage}[b]{0.48\textwidth}
        \centering
        \subfigure[]{\includegraphics[width=\linewidth]{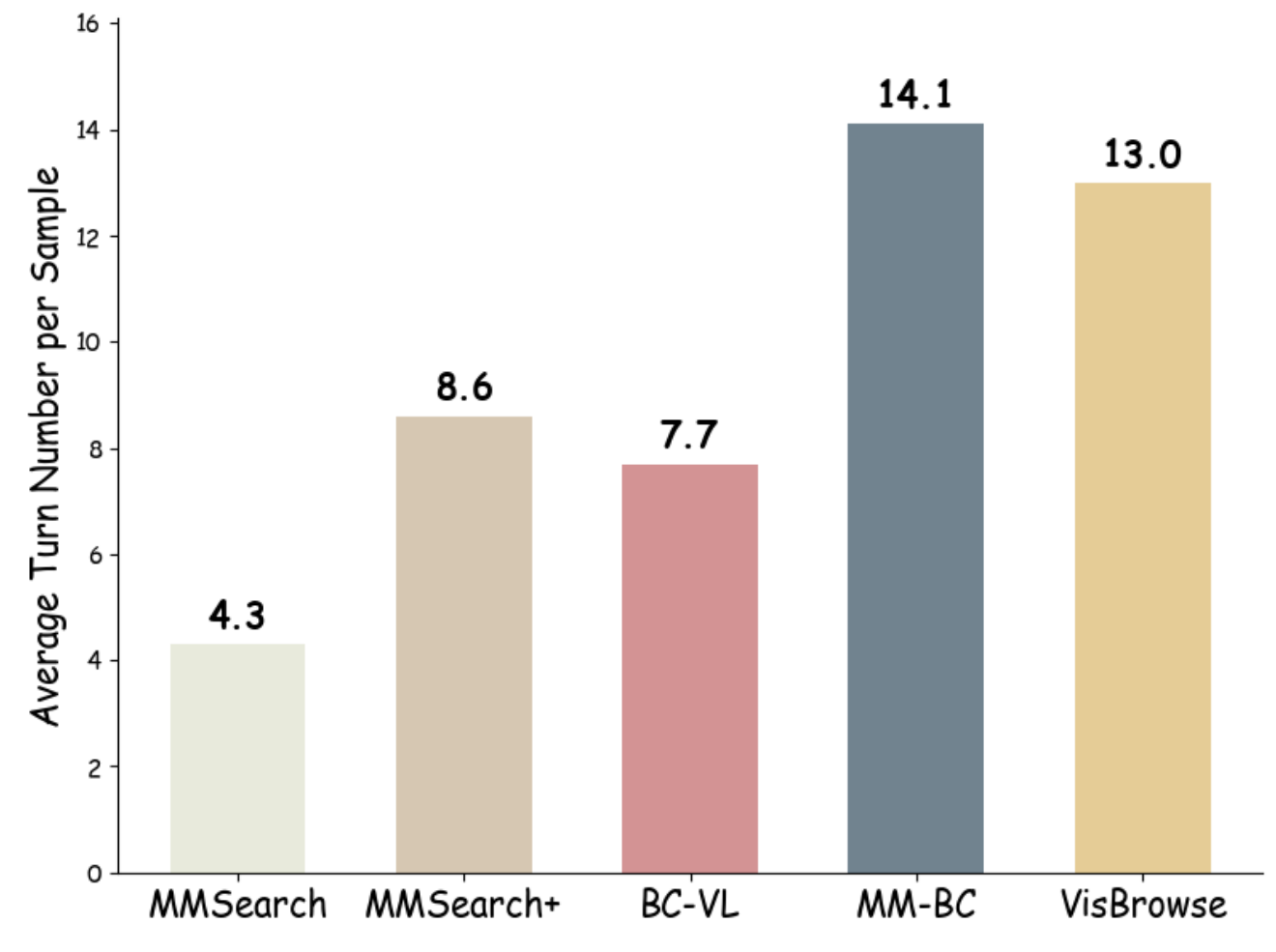}}
    \end{minipage}
    \begin{minipage}[b]{0.5\textwidth}
        \begin{subfigure}[]{
            \begin{tabular}{lccccc}
            \toprule
            \textbf{Benchmarks} & \textbf{TS} & \textbf{RIS} & \textbf{SI} & \textbf{IC} & \textbf{Visit} \\
            \midrule
            \multicolumn{1}{l|}{MMSearch} & 51.4 & 8.5 & 1.3 & 1.9 & 36.9 \\
            \multicolumn{1}{l|}{MMSearch+} & 58.4 & 13.3 & 1.3 & 6.6 & 20.4 \\
            \multicolumn{1}{l|}{BC-VL} & 64.7 & 7.5 & 0.6 & 1.7 & 25.6 \\
            \multicolumn{1}{l|}{MM-BC} & 62.8 & 6.5 & 3.2 & 2.2 & 25.3 \\
            \multicolumn{1}{l|}{VisBrowse} & 48.2 & 15.0 & 9.3 & 8.4 & 19.1 \\
            \bottomrule
            \end{tabular}
            }
        \end{subfigure}
    \end{minipage}
    \caption{(a) Average number of turns of tool interactions required per sample across the five benchmarks. (b) Distribution (\%) of different tool types across five benchmarks.}
    \label{fig5}
\end{figure}

As shown in Figure \ref{fig5}(a), we statistically analyze the average number of rounds the model interacted with the tool across five benchmarks. For relatively simple benchmarks, such as MMSearch, our model's average number of interaction turns is only 4.3. And for more challenging benchmarks, such as MM-BrowseComp, the tool's interaction turns increase to 14.1. As shown in Figure \ref{fig5}(b), to analyze the patterns of tool usgae, we calculate the distribution of different tools. Across all benchmarks, the model tends to invoke \texttt{text\_search} tool because textual evidence dominates the search trajectory for each benchmark. Compared to other benchmarks that only require a single inverse image search to obtain the semantics of an image query, VisBrowse requires more \texttt{reverse\_image\_search} and \texttt{search\_image} tool calls. This indicates that the benchmark relies on obtaining visual evidence from web pages. The case study of our search trajectory can be found in Appendix \ref{ap:case}. 

\section{Conclusion}

In this paper, we formalize the limitations of existing search agents, text-only systems suffer from visual blindness, while multimodal extensions treat vision as a passive input. To address these limitation, we propose \textbf{Visual-Seeker}, a visual-native multimodal deep search agent that unifies fine-grained visual entity perception with active visual evidence harvesting across multi-hop trajectories. We further design a active visual reasoning data synthesis pipeline that extracts complex entities from multi-entity real-world images and strategically injects visual evidence, yielding 5K high-quality training trajectories. Our agent learns active visual reasoning capbaility from these data, achieving the state-of-the-art performance across five challenging benchmarks, particularly in scenarios demanding precise multi-entity grounding and cross-modal evidence integration.

\bibliographystyle{unsrtnat} 
\bibliography{main}

\newpage
\appendix
\section{Appendix}
\label{sec:appendix}

\subsection{Data Example}
\label{ap:data}

\begin{figure*}[h]
    \centering
    \includegraphics[width=1\linewidth]{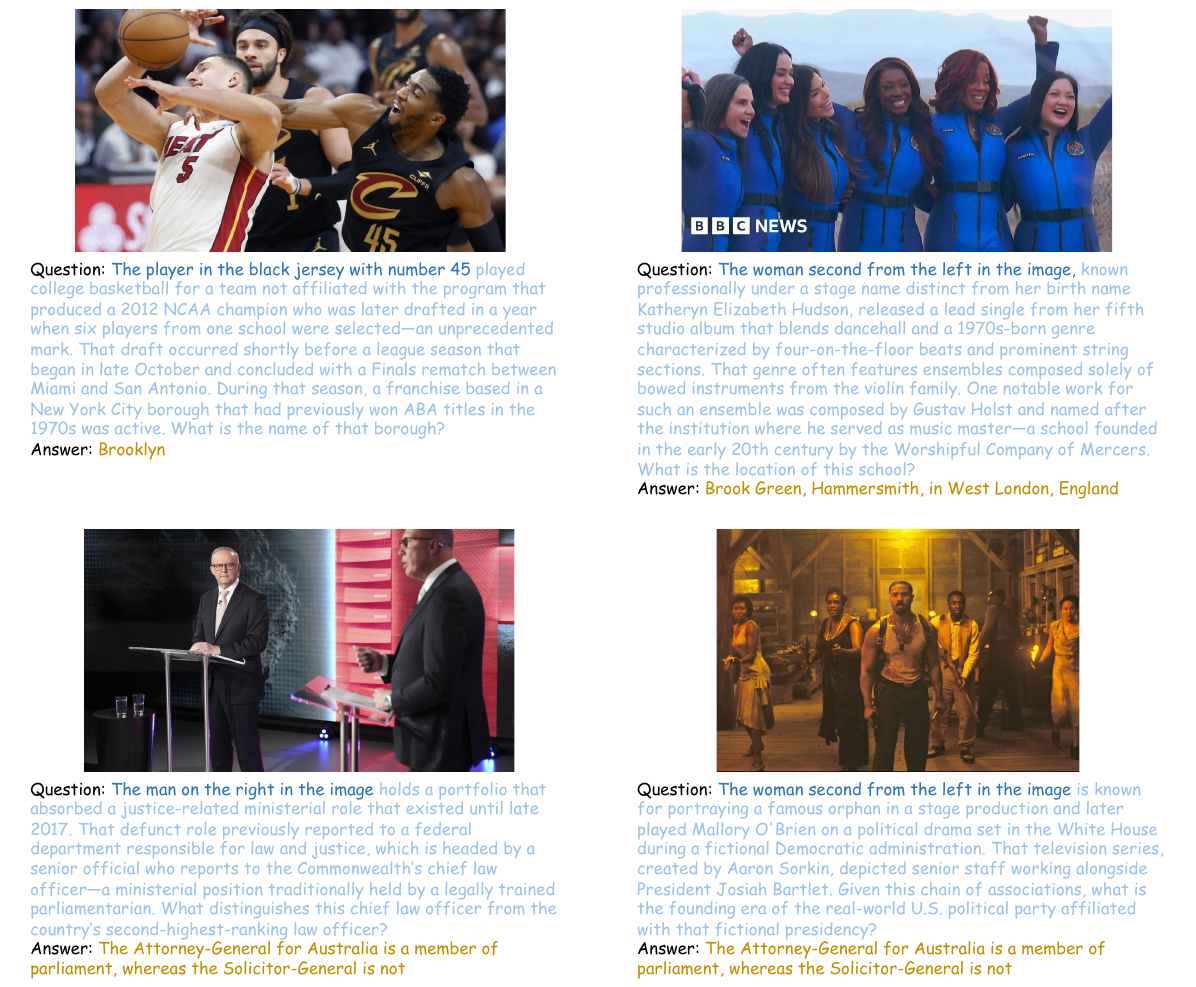}
    \caption{Data examples without visual evidence injection.}
    \label{fig6}
\end{figure*}

\begin{figure*}[h]
    \centering
    \includegraphics[width=1\linewidth]{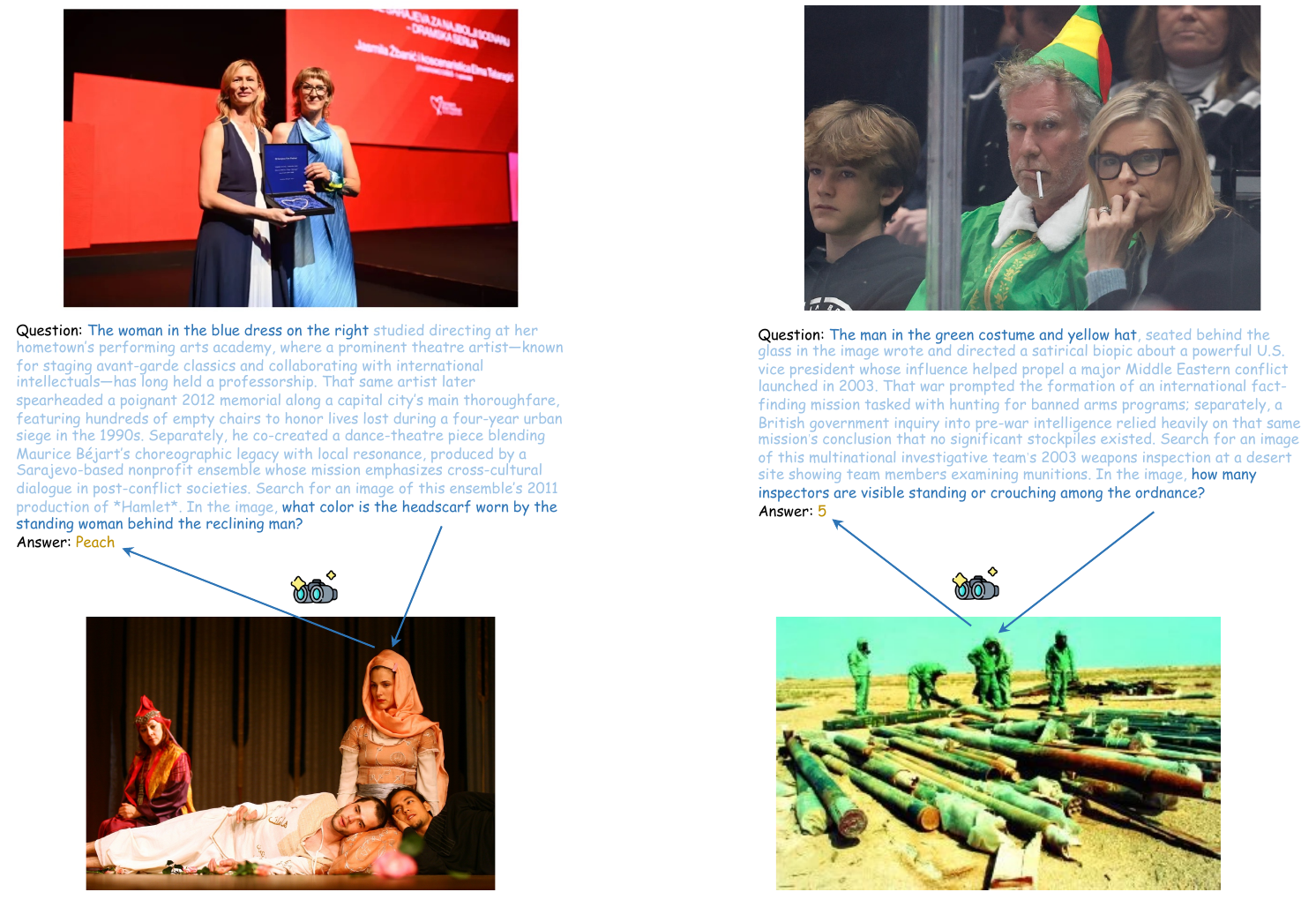}
    \caption{Data examples with visual evidence injection.}
    \label{fig7}
\end{figure*}

Based on our data synthesis pipeline, we synthesized a 5K multi-hop VQA dataset containing complex entity queries and visual evidence. Figure \ref{fig6} shows a data example without visual evidence injection, and Figure \ref{fig7} shows a data example with visual evidence injection.

\subsection{Prompt}
\label{ap:prompt}

\begin{tcolorbox}[title=System Prompt]
You are a Web Information Seeking Master. Your task is to thoroughly seek the internet for information and provide accurate answers to visual questions.

\vspace{3pt}

As you proceed, adhere to the following principles:

\vspace{3pt}

1. Decompose the original visual question into sub-questions and solve them step by step. Summarize the knowledge obtained from the previous round of dialogue, then think about what is next sub-question.

\vspace{3pt}

2. Whether you can answer the question or not, you should describe the image in detail. if the image includes multiple sub-image, you should describe each one separately.

\vspace{3pt}

3. Before calling any tools, you must provide a brief explanation of why you are calling the tool and what you expect to achieve.

\vspace{3pt}

4. You should provide the final answer within 15 turns, regardless of whether all valid information has been collected.
\end{tcolorbox}

\begin{tcolorbox}[title=Input Prompt]
You are an intelligent agent engaged in a conversation with a user. The user poses a question and provides a corresponding image for context. As an agent, you approach the problem with care and methodical precision, following a multi-step process to arrive at a solution. You utilize a variety of tools, ensuring that the information gathered from each one is cross-validated before you reach a final answer. Rather than relying on any single tool for accuracy, you employ multiple tools iteratively to prioritize the comprehensiveness and reliability of your responses.

\vspace{3pt}

To be successful, it is very important to follow the following rules:

\vspace{3pt}

1. The assistant starts with one or more cycles of (thinking about which tool to use -> performing tool call -> waiting for tool response), and ends with (thinking about the answer -> answer of the question).

\vspace{3pt}

2. If additional visual information is needed during the search process, 'search\_image' tool can be used to search for images.

\vspace{3pt}

3. You can use 'image\_crop' tool to zoom in on a specific region of the image and search for it.

\vspace{3pt}

4. You can only make one tool call per round and wait for the tool's response.

\vspace{3pt}

5. Your answer should be inside '<answer></answer>' tags, and the answer must be the most concise output.

\vspace{3pt}

Input Question:

Input image\_url:

\end{tcolorbox}

\begin{tcolorbox}[title=LLM-as-Judge Prompt]
You are a rigorous question-and-answer quality assessment expert. Given a question, your task is to compare the "groud truth" and the "model answer", determine whether they are semantically equivalent, and output "yes" or "no".

\vspace{3pt}

question: {}

ground truth: {}

model answer: {}

\end{tcolorbox}

\subsection{Case Study}
\label{ap:case}

The tasks in VisBrowse-Bench include both fine-grained entity extraction and visual evidence collection. Therefore, we conduct a case study on this benchmark to analyze the visual-native ability of our model.

\paragraph{Question:} The person in the picture is wearing a necklace from a certain brand. In 2018, a documentary about the founder of that brand was released. What fruit is the protagonist eating in the documentary poster?

\paragraph{Ground Truth:} Banana

\begin{figure}[h]
    \centering
    \includegraphics[width=0.4\linewidth]{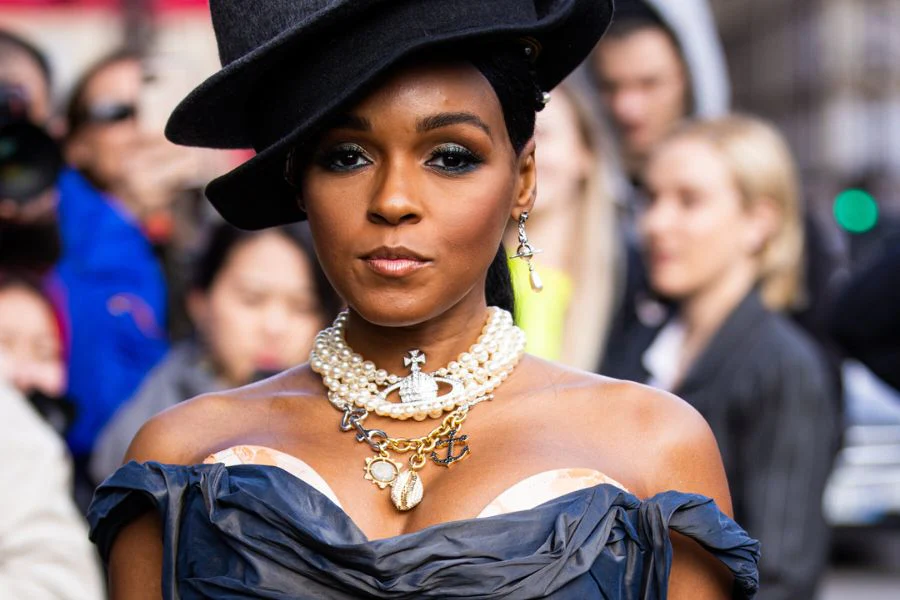}
    \caption{Visual query}
    \label{fig:placeholder}
\end{figure}

\definecolor{myblue}{RGB}{68,114,194}

\definecolor{mybrown}{RGB}{127,96,0}

\begin{tcolorbox}[title=Trajectory in VisBrowse-Bench]

\colorbox{gray!60}{\parbox{\linewidth}{\centering \textbf{Step 1}}}

\vspace{1em}

\textbf{Assistant:}

I'll analyze this step by step. Let me first examine the image closely to identify the necklace brand.

\textcolor{myblue}{\textit{<tool\_call>}} \\
\textcolor{mybrown}{image\_crop ([0.25, 0.35, 0.75, 0.75])} \\
\textcolor{myblue}{\textit{</tool\_call>}}

\textbf{Tool:}

The url of cropped image: https://...

\vspace{1em}

\colorbox{gray!60}{\parbox{\linewidth}{\centering \textbf{Step 2}}}

\vspace{1em}

\textbf{Assistant:}

I can see a pearl necklace with what appears to be a Vivienne Westwood Orb logo pendant. Let me confirm this by searching the image.

\textcolor{myblue}{\textit{<tool\_call>}} \\
\textcolor{mybrown}{reverse\_image\_search ("http://...")} \\
\textcolor{myblue}{\textit{</tool\_call>}}

\textbf{Tool:}

The title of webpage: Starship Pearl Beaded Statement Choker Necklace - Silver | Fashion Nova; 

The url of image: https://...; 

The url of webpage: https://...

...

\vspace{1em}

\colorbox{gray!60}{\parbox{\linewidth}{\centering \textbf{Step 3}}}

\vspace{1em}

\textbf{Assistant:}

The necklace appears to be a Vivienne Westwood piece. The orb/crown logo is clearly visible on the pearl choker. Now let me search for the documentary about Vivienne Westwood's founder in 2018.

\textcolor{myblue}{\textit{<tool\_call>}} \\
\textcolor{mybrown}{text\_search ("documentary 2018 Vivienne Westwood founder Alexander McQueen")} \\
\textcolor{myblue}{\textit{</tool\_call>}}

\end{tcolorbox}

\begin{tcolorbox}

\textbf{Tool:}

A Google search for 'documentary 2018 Vivienne Westwood founder Alexander McQueen' found 5 results:

Web Results

1. [Watch an exclusive clip for the new Vivienne Westwood documentary](https://www...)

Date published: Apr 30, 2018

Visually, the film will be a mixture of archival footage, reconstructions and interviews with Westwood's extensive network of collaborators.

...

\vspace{1em}

\colorbox{gray!60}{\parbox{\linewidth}{\centering \textbf{Step 4}}}

\vspace{1em}

\textbf{Assistant:}

I see two documentaries from 2018: "Westwood: Punk, Icon, Activist" about Vivienne Westwood, and "McQueen" about Alexander McQueen. The question says "a necklace from a certain brand" and then asks about "a documentary about the founder of that brand." 

The necklace is Vivienne Westwood, and Vivienne Westwood is a brand. But wait - the question could also refer to Alexander McQueen since McQueen worked closely with Vivienne Westwood. Let me re-read: "a documentary about the founder of that brand." Vivienne Westwood is the founder of Vivienne Westwood. The documentary is "Westwood: Punk, Icon, Activist" (2018).

Let me search for the poster of this documentary to see what fruit the protagonist is eating.

\textcolor{myblue}{\textit{<tool\_call>}} \\
\textcolor{mybrown}{search\_image ("Westwood Punk Icon Activist documentary 2018 poster")} \\
\textcolor{myblue}{\textit{</tool\_call>}}

\textbf{Tool:}

The title of webpage: Westwood: Punk, Icon, Activist (2018) - IMDb

The url of image: https://m.media...

The url of webpage: https://www.imdb...

...

\vspace{1em}

\colorbox{gray!60}{\parbox{\linewidth}{\centering \textbf{Step 5}}}

\vspace{1em}

\textbf{Assistant:}

I found the documentary poster for "West wood: Punk, Icon, Activist" (2018). In the poster, Vivienne Westwood (the protagonist) is clearly shown eating a **banana**. The Japanese poster even has text that reads "Vivienne eating banana."

Let me verify this by looking at the poster image more closely.

\textcolor{myblue}{\textit{<tool\_call>}} \\
\textcolor{mybrown}{visit(goal="Find the poster of the documentary and identify what fruit Vivienne Westwood is eating", url="https://www.imdb.com/...")} \\
\textcolor{myblue}{\textit{</tool\_call>}}

\textbf{Tool:}

The useful information in https://www.imdb.com... for user goal Find the poster of the documentary and identify what fruit Vivienne Westwood is eating as follows: 

Evidence in page: 

\end{tcolorbox}

\begin{tcolorbox}

...

Summary: 

...

\vspace{1em}

\colorbox{gray!60}{\parbox{\linewidth}{\centering \textbf{Step 6}}}

\vspace{1em}

\textbf{Assistant:}

I need to find more usefule information about Westwood

\textcolor{myblue}{\textit{<tool\_call>}} \\
\textcolor{mybrown}{reverse\_image\_search ("http://...")} \\
\textcolor{myblue}{\textit{</tool\_call>}}

\textbf{Tool:}

The title of webpage: Amazon.com: Westwood: Punk, Icon, Activist : Vivienne Westwood, Lorna Tucker: Movies \& TV

The url of image: https://m.media...

The url of webpage: https://www...

...

\vspace{1em}

\colorbox{gray!60}{\parbox{\linewidth}{\centering \textbf{Step 7}}}

\vspace{1em}

\textbf{Assistant:}

The evidence is clear. In the documentary poster for "Westwood: Punk, Icon, Activist" (2018), Vivienne Westwood is shown eating a **banana**. The Japanese poster even includes text confirming "Vivienne eating banana."

\textcolor{green!70!black}{<answer>Banana</answer>}

\end{tcolorbox}

\end{document}